\documentclass[runningheads]{llncs}

\usepackage{eccv}

\usepackage{eccvabbrv}
\usepackage{algorithm}
\usepackage{algpseudocode}
\usepackage{graphicx}
\usepackage{booktabs}
\usepackage{tabularray}
\usepackage[accsupp]{axessibility}  
\usepackage[colorlinks = true,
            urlcolor  = magenta,
            citecolor = black,
            ]{hyperref}

\usepackage{hyperref}

\usepackage{orcidlink}
\makeatletter
\newcommand{\printfnsymbol}[1]{%
  \textsuperscript{\@fnsymbol{#1}}%
}
\makeatother

\begin{document}

\title{DQ-DETR: DETR with Dynamic Query for Tiny Object Detection}
\titlerunning{DQ-DETR: DETR with Dynamic Query for Tiny Object Detection}

\author{Yi-Xin Huang \inst{1}\thanks{Equal contribution} \and
Hou-I Liu\printfnsymbol{1} \inst{1}\and
Hong-Han Shuai\inst{1} \and
Wen-Huang Cheng\inst{2}}

\authorrunning{Y.~Huang et al.}

\institute{National Yang Ming Chiao Tung University, Hsinchu, Taiwan
\email{\{svkatie.nctu.ee08, k39967.c, hhshuai\}@nycu.edu.tw} \and
National Taiwan University, Taipei, Taiwan \\
\email{wenhuang@csie.ntu.edu.tw}}




\maketitle
\begin{abstract}
Despite previous DETR-like methods having performed successfully in generic object detection, tiny object detection is still a challenging task for them since the positional information of object queries is not customized for detecting tiny objects, whose scale is extraordinarily smaller than general objects. 
Additionally, the fixed number of queries used in DETR-like methods makes them unsuitable for detection if the number of instances is imbalanced between different images. 
Thus, we present a simple yet effective model, DQ-DETR, consisting of three components: categorical counting module, counting-guided feature enhancement, and dynamic query selection to solve the above-mentioned problems. 
DQ-DETR uses the prediction and density maps from the categorical counting module to dynamically adjust the number and positional information of object queries. Our model DQ-DETR outperforms previous CNN-based and DETR-like methods, achieving state-of-the-art mAP 30.2\% on the AI-TOD-V2 dataset, which mostly consists of tiny objects.
Our code will be available at \url{https://github.com/hoiliu-0801/DQ-DETR}.
\end{abstract}    
\section{Introduction}
\label{sec:intro}

\hspace{\parindent}Convolutional neural networks (CNNs) excel at processing the RGB semantic and spatial texture features. 
Most object detection methods are primarily based on CNNs.
For example, Faster R-CNN~\cite{frcnn} introduces a region proposal network to generate potential object regions.
FCOS~\cite{fcos} applies a center prediction branch to increase the quality of the bounding boxes.

However, CNNs are unsuitable for capturing long-range dependencies in the image, restricting the detection performance.
Recently, DETR~\cite{DETR} incorporates CNN and transformer architecture to establish a new object detection framework. DETR utilizes the transformer encoder to integrate the partitioned image patches and passes them with the learnable object queries to the transformer decoder for final detection results. Moreover, a series of DETR-like methods~\cite{Deformable-DETR,DN-DETR,DINO,DAB-DETR} aim to advance DETR performance and accelerate convergence speed. For example, Deformable-DETR~\cite{Deformable-DETR} uses multi-scale feature maps to improve its ability to detect different sizes of objects. Also, the use of deformable attention modules can not only capture more informative and contextually relevant features but accelerate training convergence as well.

\begin{table}
\caption{Comparison of DETR-like models' query strategies under different situations.}
\centering
\resizebox{0.999\linewidth}{!}{
\begin{tblr}{
  hlines,
  vline{2-5} = {-}{},
  column{2-4} = {c},
  column{5} = {c}, 
}
                  & Sparse & Dense & Imbalance & Characteristics                                                     \\
Deformable
  DETR~\cite{Deformable-DETR} & \checkmark      &       &           & Sparse Queries
  (K=300) with One-To-One Assignment; Low Recall~    \\
DDQ-DETR~\cite{DDQ}          & \checkmark      & \checkmark     &           & Dense
  Distinct Queries (K=900); Low Recall if \#Object $\gg$ \#Query \\
DQ-DETR (Ours)     & \checkmark      & \checkmark     & \checkmark         & Dynamically
  adjust the "Number" and "Position" of Queries~        
\end{tblr}
}
\label{intro}
\end{table}

In this work, we argue that the previous DETR-like methods are inappropriate in aerial image datasets, which only contain tiny objects and have an imbalance of instances between different images. In the previous DETR-like methods, the object queries used in the transformer decoder do not consider the number and position of instances in the image. Generally, they apply a fixed number K of object queries, where K represents the maximum number of the detection objects, e.g., K=100, 300 in DETR and Deformable-DETR, respectively. 
DETR~\cite{DETR} and Deformable-DETR~\cite{Deformable-DETR} apply a fixed number of sparse queries, suffering from a low recall rate. To address this problem, DDQ~\cite{DDQ} selects dense distinct queries, K=900, with a class-agnostic NMS based on a hand-designed IoU threshold. Though DDQ applies dense queries for detection, the number of queries is still limited. 

However, aerial datasets often exhibit imbalances in the distribution of instances across different images.
A fixed number of queries can lead to poor detection accuracy when the number of objects varies drastically between images.
For example, in the AI-TOD-V2 dataset~\cite{AITODv2}, some images have more than 1500 objects, but others have less than 10 objects. Under the situation that the number of objects in images is more than DETR's query number K, a low recall rate is an expected issue. 
Using smaller K restricts the recall of the objects in dense images, leaving many instances undetected (FN).
Conversely, using a large K in the sparse images not only introduces many underlying false positive samples (FP) but also causes a waste of computing resources since the computing complexity in the decoder’s self-attention layers grows quadratic with the number of queries K. 

Furthermore, in the previous DETR-like methods, the object queries do not consider the position of instances in the image. 
The position of object queries is a set of learned embeddings, which are irrelevant to the current image and do not have explicit physical meaning to tell where the queries are focusing on.  
The static positions of object queries are unsuitable for aerial image datasets, where the distribution of instances varies extremely in different images, i.e., some images contain dense objects concentrated in specific areas, while some only have a few objects scattered throughout the images.

Stemming from the above-mentioned weakness, we propose a novel DETR-like method named DQ-DETR, which mainly focuses on dynamically adapting the numbers of queries and enhancing the position of queries to locate the tiny objects precisely.
In this study, we propose a dynamic query selection module for adaptively choosing different numbers of object queries in DETR's decoder stage, resulting in fewer FP in sparse images and fewer FN in dense images. 
Moreover, we generate the density maps and estimate the number of instances in an image by the categorical counting module. The number of object queries is adjusted based on the predicted counting number.
In addition, we aggregate the density maps with the visual feature from the transformer encoder to reinforce the foreground features, enhancing the spatial information for tiny objects.
The strengthened visual feature will be further used to improve the positional information of object queries.
As such, we can simultaneously handle the images with few and crowded tiny objects by dynamically adjusting the number and position of object queries used in the decoder.

Our contributions are summarized as follows:
\begin{itemize}
    \item We point out the crucial limitation of previous DETR-like methods that make them unsuitable for aerial image datasets.
    \item We propose three components: the categorical counting module, counting-guided feature enhancement, and dynamic query selection. These components significantly enhance performance on tiny objects.
    \item Experimental result shows that our proposed DQ-DETR significantly surpasses the state-of-the-art method by 16.6\%, 20.5\% in terms of AP, $\text{AP}_{vt}$ on the AI-TOD-V2 dataset.
\end{itemize}
\section{Related work}
\label{sec:related work}

\subsection{Tiny Object Detection}
Detecting small objects is challenging due to their lack of pixels.
Early works apply data augmentation to oversample the instance of tiny objects. For example,~\cite{da_sm}, copy-paste small objects into the same image.~\cite{learn_da} proposes a K sub-policies that automatically transform features from the instance level. 
In addition, several approaches, such as~\cite{AITODv2,nwd,rfla,dot}, indicate that traditional Intersection over Union (IoU) metrics are ill-suited for tiny objects.
When the object size difference is significant, IoU becomes highly sensitive.
To design the appropriate metrics for the tiny object, DotD~\cite{dot} considers the object's absolute and relative size to formulate a new loss function.~\cite{nwd,rfla,dot} design a new label assignment based on Gaussian distribution, which alleviates the sensitivity of the object size. However, these methods heavily rely on the predefined threshold, which is unstable for a different dataset.

\subsection{DETR-like Methods}
DETR~\cite{DETR} proposes an end-to-end object detection framework based on the transformer, where the transformer encoder extracts instance-level features from an image,  and the transformer decoder uses a set of learnable queries to probe and pool features from images.
While DETR achieves comparable results with the previous classical CNN-based detectors~\cite{frcnn,fcos}, it suffers severely from the problem of slow training convergence, needing 500 epochs of training to perform well. Many follow-up works have attempted to address the slow training convergence of DETR from different perspectives. 

Some argue that DETR's slow convergence stems from the instability of Hungarian matching and the cross-attention mechanism in the transformer decoder.~\cite{Sun_2021_ICCV} proposes an encoder-only DETR, discarding the transformer decoder. Dynamic DETR~\cite{Dynamic_DETR} designs an ROI-based dynamic attention mechanism in the decoder that can focus on regions of interest from a coarse-to-fine manner. Deformable-DETR~\cite{Deformable-DETR} proposes an attention module that only attends to a few sampling points around a reference point. DN-DETR~\cite{DN-DETR} introduces denoising training to reduce the difficulty of bipartite graph matching.

Another series of works makes improvements in decoder object queries. Since the object queries are just a set of learnable embedding in DETR,~\cite{Anchor_DETR, Conditional_DETR, DAB-DETR} imputes the slow convergence of DETR to the implicit physical explanation of object queries. Conditional DETR~\cite{Conditional_DETR} decouples the decoder’s cross-attention formulation and generates conditional queries based on reference coordinates. DAB-DETR~\cite{DAB-DETR} formulates the positional information of object queries as 4-D anchor boxes $(x, y, w, h)$ that are used to provide RoI (Region of Interest) information for probing and pooling features. Although DETR-like methods have improved the formulation of queries, they are constrained in their ability to handle tiny objects and datasets with widely varying numbers of objects. The object queries in these methods are learned from the training data and the number of queries remains the same across different input images. 
%

 rmore, while DETR-like methods have improved the formulation of queries, they are constrained in their ability to handle tiny objects and datasets with widely varying numbers of objects.
Our proposed DQ-DETR stands out as the first DETR-like model specifically designed to detect tiny objects and dynamically adjust the number of queries to enhance precision in imbalanced datasets.
\section{Method}
\begin{figure*}[ht!]
    \centering
    \includegraphics[width=1\linewidth]{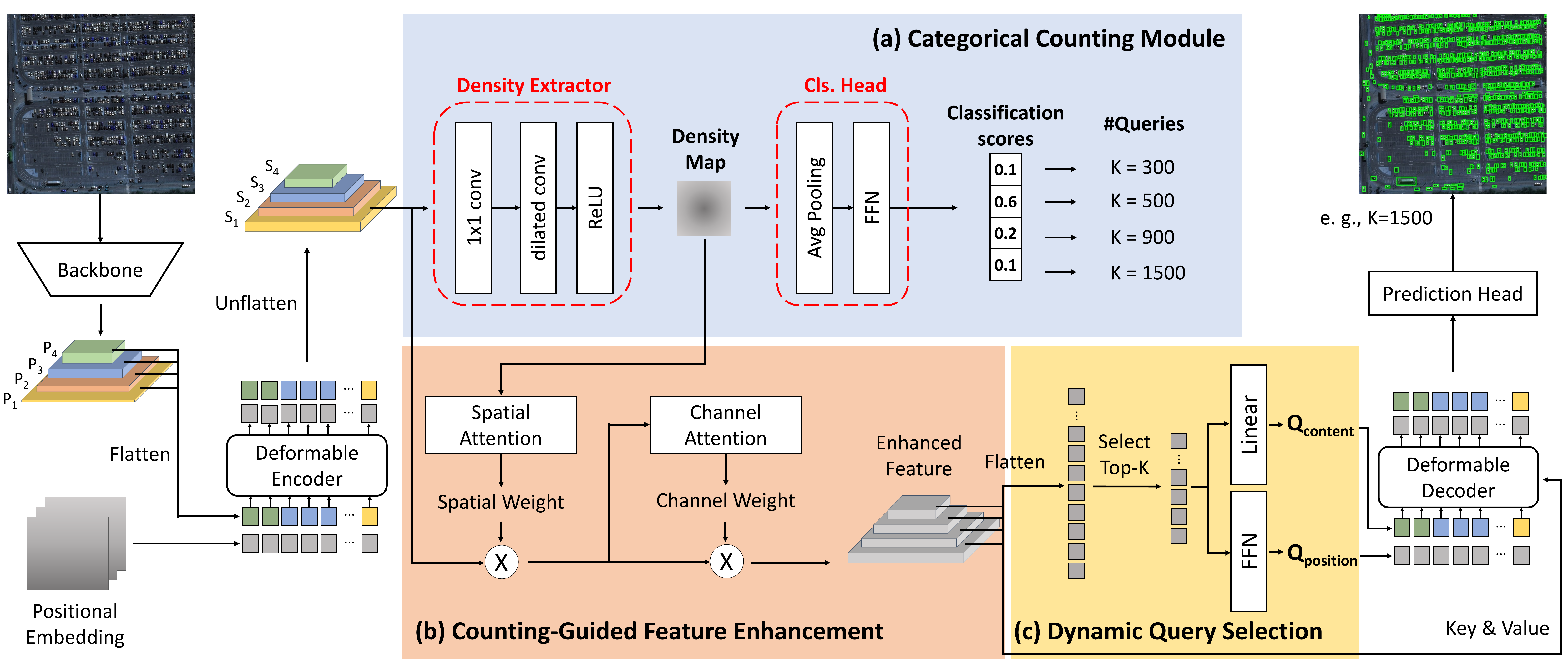}
    \caption{The overall architecture of our method. (a) Categorical Counting Module, which classifies the number of instances in images into 4 levels. (b) Counting-Guided Feature Enhancement, which refines the encoder's visual feature with a density map. (c) Dynamic Query Selection, which dynamically adjusts the number of queries and enhances the content and position of queries.}
    \label{fig:model_method}
\end{figure*}

\subsection{Overview} 
\hspace{\parindent}The overall structure of DQ-DETR is shown in Fig. \ref{fig:model_method}.
As a DETR-like method, DQ-DETR is an end-to-end detector that contains a CNN backbone, a deformable encoder and decoder~\cite{Deformable-DETR}, and several prediction heads.
We further implement a new categorical counting module, a counting-guided feature enhancement module, and dynamic query selection based on DETR's architecture. 
Given an input image, we first extract multi-scale features with a CNN backbone and feed them into the transformer encoder to attain visual features. 
Afterward, our categorical counting module determines how many object queries are used in the transformer decoder, as shown in Fig.~\ref{fig:model_method}(a).
Besides, we propose a novel counting-guided feature enhancement module, as illustrated in Fig.~\ref{fig:model_method}(b), to strengthen the encoder's visual features with spatial information for tiny objects. 
Last, the object queries are refined with additional information about the location and size of tiny objects through dynamic query selection, as shown in Fig.~\ref{fig:model_method}(c).
The following section will describe the proposed Categorical Counting Module, Counting-Guided Feature Enhancement, and Dynamic Query Selection.

\subsection{Unflattening of Encoder's Feature Map}
\hspace{\parindent}Following DETR's pipeline, we use multi-scale feature maps ${P}_{i} \in \{1, 2, \dots, l\}$ extracted from different stages of the backbone as the input of the transformer encoder.
To form the input sequence of the transformer encoder, we flatten each layer of multi-scale feature maps ${P}_{i}$ from $\mathbb{R}^{d \times h_{i}\times w_{i}}$ to $\mathbb{R}^{d \times h_{i}w_{i}}$ and then concatenate them together.
The higher resolution feature contains more spatial details, which is beneficial to object counting and detecting tiny objects.

In our proposed categorical counting module, we apply dilated convolution operations on the transformer encoder features.
Hence, we unflatten the encoder's multi-scale visual features by reshaping its spatial dimension, resulting in 2-D feature maps ${S}_{i} \in \mathbb{R}^{d \times h_{i} \times w_{i}}$.
We denote the reconstructed encoder’s multi-scale visual features as EMSV features for brevity.

\subsection{Categorical Counting Module}
\hspace{\parindent}The categorical counting module aims to estimate the number of objects in the images. It consists of a density extractor and a classification head. 
\subsubsection{Density Extractor.}
We take the largest feature map $S_{1}$ of the EMSV features and generate the density map ${F}_{c}$ through the density extractor. High-resolution features are essential for detecting tiny objects, as they provide a clearer representation of such objects.
The input feature map $S_{1}$ is sent into a series of dilated convolution layers to acquire a density map ${F}_{c}$, which contains counting-related information.  
Specifically, dilated convolution layers enlarge the receptive field and capture rich long-range dependency for tiny objects.

\subsubsection{Counting Number Classification.}
\hspace{\parindent}Lastly, we estimate the counting number $N$, i.e., the number of instances per image, by a classification head and categorize them into four levels, which are $N \leq 10$, $10 < N \leq  100$, $100 < N \leq 500$, and $N > 500$. The classification head consists of two linear layers.
Further, the numbers 10, 100, and 500 are selected based on the AI-TOD-V2 dataset's characteristics, i.e., the mean and standard deviation of the number of instances $N$ per image.
Notably, we do not use the regression head as in the traditional crowd-counting methods, which regresses the counting number to a specific number.
We attribute the reason to the drastic difference in the number of instances in each image, where $N$ ranges from 1 to 2267 in different images of AI-TOD-V2. It is difficult to regress an accurate number, hurting the detection performance. 


\subsection{Counting-Guided Feature Enhancement Module (CGFE)}
\hspace{\parindent} 
The EMSV feature is refined using the density map from the categorical counting module through the proposed Counting-Guided Feature Enhancement Module (CGFE) to improve the spatial information of tiny objects.
The refined features are then used to enhance the position information of queries.
This module comprises spatial cross-attention and channel attention operations~\cite{CBAM}. 

\subsubsection{Spatial cross-attention map.}
To utilize the abundant spatial information in the density map ${F}_{c}$, a 2-D cross-spatial attention is calculated.
We employ a $1 \times 1$ convolution layers to down-sample the density map ${F}_{c}$, creating multi-scale counting feature maps ${F}_{c, i} \in \{1, 2, \dots, l\}$ to in line with the shape of each layer of encoder’s multi-scale feature maps $S_{i} \in \{1, 2, \dots, l\}$.
Subsequently, we first apply average pooling (AvgP.) and max pooling (MaxP.) on each layer of multi-scale counting features ${F}_{c, i} \in \mathbb{R}^{b\times 256\times h\times w}$ along the channel axis. 
Then, the two pooling features $\mathbb{R}^{b \times 1 \times h\times w}$ are concatenated and sent into a 7x7 convolution layer followed by a sigmoid function to produce spatial attention map ${W}_{s} \in \mathbb{R}^{b \times 1 \times h\times w}$. We formulate this process in Eq.~\ref{eq:sp_map}. 

Since the density maps ${F}_{c}$ contain the location and density information about the object, the spatial attention maps generated by them can focus on the important region, i.e., foreground objects, and enhance the EMSV feature with abundant spatial information. 
\noindent
\begin{equation}
\begin{aligned}
{W}_{s, i} = \sigma(\underset{7 \times 7}{Conv}(Concat\begin{bmatrix}
  AvgP.(\underset{1 \times 1}{Conv}({F}_{c, i})) \\
MaxP.(\underset{1 \times 1}{Conv}({F}_{c, i}))
\end{bmatrix}
))
\end{aligned}
\label{eq:sp_map}
.
\end{equation}

The generated spatial attention map ${W}_{s, i}$ multiplies with EMSV feature $S_{i}$ element-wisely and further obtains the spatial-intensified features ${E}_{i}$, as shown in Eq.~\ref{eq:ei}.
\noindent
\begin{equation}
\begin{aligned}
{E}_{i} = W_{s, i} \otimes S_{i},
\end{aligned}
\label{eq:ei}
\end{equation}

\subsubsection{Channel attention map.}
After the spatial cross-attention, we further apply 1-D channel attention to the spatial-intensified features ${E}_{i}$, exploiting the inter-channel relationship of features.
Specifically, we first apply average pooling and max pooling on each layer of ${E}_{i} \in \mathbb{R}^{b\times 256\times h\times w}$ along the spatial dimension. 
Next, the two pooling features $\mathbb{R}^{b \times 256 \times 1 \times 1}$ are sent into a shared MLP and merged together with element-wise addition to create channel attention map ${W}_{c, i}$.
Finally, the channel attention map ${W}_{c, i} \in \mathbb{R}^{b \times 256 \times 1 \times 1}$ is multiplied with original $E_{i} \in \mathbb{R}^{b \times 256 \times h \times w}$ and further get the counting-guided intensified feature maps ${F}_{t}$. The formulas are defined in Eq.~\ref{eq:wci} and Eq.~\ref{eq:fti}: 
\noindent
\begin{equation}
\begin{aligned}
{W}_{c, i} = \sigma(MLP(AvgP.({E}_{i})) + MLP(MaxP.({E}_{i}))),
\end{aligned}
\label{eq:wci}
\end{equation}
\noindent
\vspace{-5pt}
\begin{equation}
\begin{aligned}
{F}_{t, i} = W_{c, i} \otimes E_{i}.
\end{aligned}
\label{eq:fti}
\end{equation}

\subsection{Dynamic Query Selection}
\subsubsection{Number of queries.}
In dynamic query selection, we first use the classification result from the categorical counting module to determine the number of queries $K$ used in the transformer decoder. 
The four classification classes in the categorical counting module correspond to four distinct numbers of queries, which are $K$ = 300, 500, 900, and 1500, i.e., if the image is classified as $N \leq 10$, we use $K=300$ queries in the subsequent detection task, and so forth.

\subsubsection{Enhancement of queries.}
For query formulation, we follow the idea of DAB-DETR~\cite{DAB-DETR}, where the queries are composed of content and positional information. The content of queries is a high-dimension vector, while the position of queries is formulated as a 4-D anchor box (x, y, w, h) to accelerate training convergence. 

Further, we use the intensified multi-scale feature maps ${F}_{t}$ from the previous CGFE module to improve the content ${Q}_{content}$ and position ${Q}_{position}$ of queries. 
Each layer of ${F}_{t}$ is firstly flattened into pixel level and concatenated together, forming ${F}_{flat} \in \mathbb{R}^{b\times 256\times hw}$.
The top-K features are selected as priors to enhance decoder queries, where $K$ is the number of queries used in the transformer decoder stage.  
The selection is based on the classification score. We feed ${F}_{flat}$ into an FFN for the object classification task and generate the classification score $\in \mathbb{R}^{b\times m\times hw}$, where m is the number of object classes in the dataset.
Consequently, we generate the content and position of queries using the selected top-K features ${F}_{select}$.
\noindent
\begin{equation}
\begin{aligned}
& Score = FFN({F}_{flat}), \\
& {F}_{select} = topK_{Score}({F}_{flat}). \\
\end{aligned}
\label{eq:dynamic_select1}
\end{equation}

The content of queries is generated by a linear transform of the selected features ${F}_{select}$.
As for the position of queries, we use an FFN to predict bias $\hat{b_{i}} = (\Delta b_{ix}, \Delta b_{iy}, \Delta b_{iw}, \Delta b_{ih})$ to refine the original anchor boxes. 
Let $(x, y)_{i}$ index a selected feature from multi-level features $F_{t} \in \{1, 2, \dots, l\}$ at position (x, y).
The selected feature has its original anchor box $(x_{i}, y_{i}, w_{i}, h_{i})$ as the position prior of queries, where $(x_{i}, y_{i}) $ are normalized coordinates $\in {[0, 1]}^2$ and $(w_{i}, h_{i}) $ are setting related to the scale of feature $F_{t}$.
The predicted bias $\hat{b_{i}} = (\Delta b_{ix}, \Delta b_{iy}, \Delta b_{iw}, \Delta b_{ih})$ are then added to original anchor box to refine the position of object queries.
\noindent
\begin{equation}
\begin{aligned}
& Q_{content} = linear({F}_{select}), \\
& Q_{position, bias} = FFN({F}_{select}). \\
\end{aligned}
\label{eq:dynamic_select2}
\end{equation}

Since the features ${F}_{select}$ are selected from ${F}_{t}$, which is generated from the previous CGFE module, they contain abundant scale and location information of tiny objects.
Hence, the enhanced content and position of object queries are tailored based on each image's density (crowded or sparse), facilitating easier localization of tiny objects in the transformer decoder.
\subsection{Overall Objective}

\subsubsection{Hungarian Loss}
Based on DETR~\cite{DETR}, we use a Hungarian algorithm to find an optimal bipartite matching between ground truth and prediction and optimize losses. 
The Hungarian loss consists of L1 loss and GIoU loss~\cite{GIOU} for bounding box regression and focal loss~\cite{focal_loss} with $\alpha = 0.25$, $\gamma = 2$ for classification task, which can be denoted as Eq.~\ref{eq:hugarian}. 
Follow the settings of DAB-DETR~\cite{DAB-DETR}, we use ${\lambda}_{1} = 5$, ${\lambda}_{2} = 2$, ${\lambda}_{3} = 1$ in our implementation.
\noindent
\begin{equation}
\begin{aligned}
L_{hungarian} = {\lambda}_{1}L_{1} + {\lambda}_{2}L_{GIoU} + {\lambda}_{3}L_{focal}.
\end{aligned}
\label{eq:hugarian}
\end{equation}

In addition, we use the cross-entropy loss in the categorical counting module to supervise the classification task. Further, the Hungarian loss is also applied as the auxiliary loss for each decoder stage. The overall loss can be denoted as:
\begin{equation}
\begin{aligned}
L_{total} = L_{hungarian} + L_{aux} + L_{counting}.
\end{aligned}
\label{eq:loss}
\end{equation}

\section{Experiments}

\subsection{Datasets}
\hspace{\parindent}To demonstrate the effectiveness of our model, we conduct experiments on the aerial dataset AI-TOD-V2~\cite{AITODv2}, which mostly consists of tiny objects.

\noindent \textbf{AI-TOD-V2.}
This dataset includes 28,036 aerial images with 752,745 annotated object instances. 
There are 11,214 images for the train set, 2,804 for the validation set, and 14,018 for the test set.
The average object size in AI-TOD-V2 is only 12.7 pixels, with 86\% of objects in the dataset smaller than 16 pixels, and even the largest object is no bigger than 64 pixels. 
Also, the number of objects in an image can vary enormously from 1 to 2667, where the average number of objects per image is 24.64, with a standard deviation of 63.94.

\noindent \textbf{VisDrone.}
This dataset includes 14,018 drone-shot images, with 6,471 images for the train set, 548 for the validation set, and 3,190 for the test set.
There are 10 categories, and the image resolution is 2000 $\times$ 1500 pixels.
Also, the images are diverse in a wide range of aspects, including objects (pedestrians, vehicles, bicycles, etc.) and density (sparse and crowded scenes), where the average number of objects per image is 40.7 with a standard deviation of 46.41.

\noindent \textbf{Evaluation Metric.}
We use the AP (Average Precision) metric with a max detection number of 1500 to evaluate the performance of our proposed method. 
Specifically, AP means the average value from $\text{AP}_{50}$ to $\text{AP}_{95}$ , with IoU interval of 0.05.
Moreover, $\text{AP}_{vt}$, $\text{AP}_{t}$, $\text{AP}_{s}$, and $\text{AP}_{m}$ are for very tiny, tiny, small, and medium scale evaluation in AI-TOD~\cite{AI-TOD}. 

\subsection{Implementation Details}
\hspace{\parindent}Based on the DETR-like structure, we use a 6-layer transformer encoder, a 6-layer transformer decoder with 256 as the hidden dimension, and a ResNet50 as our CNN backbone.
Furthermore, we train our model with Adam optimizer using NVIDIA 3090 GPUs. 
The batch size is set to 1 due to memory constraints.
The same random crop and scale augmentation strategies are applied following DETR~\cite{DETR}.
In addition, to minimize errors that propagate from the categorical counting module to dynamic query selection, we apply a two-stage training scheme.
We first train the categorical counting module to achieve more stable results for the number of queries in the transformer decoder.
After stabilizing the counting result, we add the counting-guided feature enhancement module into training to refine the encoder's visual features with density maps.

\begin{table*}[ht]
\caption{\textbf{Experiments on AI-TOD-V2.} All models are trained on the \textit{trainval} split and evaluated on the \textit{test} split. * denotes a re-implementation of the results.}
\centering
\resizebox{0.98\linewidth}{!}{
\begin{tblr}{
  row{2} = {c},
  row{11} = {c},
  column{even} = {c},
  column{3} = {c},
  column{5} = {c},
  column{7} = {c},
  column{9} = {c},
  cell{2}{1} = {c=9}{},
  cell{11}{1} = {c=9}{},
  hline{1-3,11-12,16-17} = {-}{},
}
Method                    & Backbone      & AP   & $\text{AP}_{50}$ & $\text{AP}_{75}$ & $\text{AP}_{vt}$ & $\text{AP}_{t}$ & $\text{AP}_{s}$ & $\text{AP}_{m}$ \\
\hline
\qquad \qquad CNN-based models    &                   &               &      &      &      &        &       &       &       \\
YOLOv3~\cite{yolov3}                        & Darknet53     & 4.1  & 14.6 & 0.9  & 1.1    & 4.8   & 7.7   & 8.0      \\
RetinaNet~\cite{focal_loss}                       & ResNet50-FPN & 8.9  & 24.2 & 4.6  & 2.7    & 8.4   & 13.1  & 20.2   \\
Faster-RCNN~\cite{frcnn}                     & ResNet50-FPN & 12.8 & 29.9 & 9.4  & 0.0      & 9.2   & 24.6  & 37.0     \\
Cascade R-CNN~\cite{crcnn}                     & ResNet50-FPN & 15.1 & 34.2 & 11.2 & 0.1    & 11.5  & 26.7  & 38.5  \\
DetectoRS~\cite{detectors}                      & ResNet50-FPN & 16.1 & 35.5 & 12.5 & 0.1    & 12.6  & 28.3  & 40.0  \\
DotD~\cite{dot}                            & ResNet50-FPN & 20.4 & 51.4 & 12.3 & 8.5    & 21.1  & 24.6  & 30.4  \\
NWD-RKA~\cite{AITODv2}                        & ResNet50-FPN & 24.7 & 57.4 & 17.1 & 9.7    & 24.2  & 29.8  & 39.3  \\
RFLA~\cite{rfla}                             & ResNet50-FPN & 25.7 & 58.9 & 18.8 & 9.2    & 25.5  & 30.2  & 40.2  \\
\hline
\qquad \qquad DETR-like models   &  &                &               &      &      &      &        &       &       &       \\
DETR-DC5*~\cite{DETR}                       & ResNet50     & 10.4 & 32.5 & 3.9  & 3.6    & 9.3   & 13.2  & 24.6  \\
Deformable-DETR*~\cite{Deformable-DETR}                   & ResNet50     & 18.9 & 50.0 & 10.5 & 6.5    & 17.6  & 25.3  & 34.4  \\
DAB-DETR*~\cite{DAB-DETR}                          & ResNet50     & 22.4 & 55.6 & 14.3 & 9.0    & 21.7  & 28.3  & 38.7  \\
DINO-DETR*~\cite{DINO}                         & ResNet50     & 25.9 & 61.3 & 17.5 & 12.7   & 25.3  & 32.0  & 39.7  \\
\textbf{DQ-DETR (Ours)}                   & ResNet50     & \textbf{30.2 (+4.3)} & \textbf{68.6} & \textbf{22.3} & \textbf{15.3}   & \textbf{30.5}  & \textbf{36.5}  & \textbf{44.6}  
\end{tblr}
}
\label{table:Overall}
\end{table*}

\subsection{Main Results}
\noindent \textbf{AI-TOD-V2.}
Table \ref{table:Overall} shows our main results on the AI-TOD-V2 test split. 
We compare the performances of our DQ-DETR with strong baselines, including both CNN-based and DETR-like methods. 
All CNN-based methods except YOLOv3 use ResNet50 with feature pyramid network (FPN)~\cite{FPN}.
Moreover, since there is no previous research on DETR-like models for tiny object detection, our DQ-DETR is the first DETR-like model that focuses on detecting tiny objects.
We re-implement a series of DETR-like models on AI-TOD-V2, and all DETR-like methods except DETR use 5-scale feature maps with deformable attention~\cite{Deformable-DETR}. 
For 5-scale feature maps, features are extracted from stages 1, 2, 3, and 4 of the backbone, and add the extra feature by down-sampling the output of stage 4.  

The results are summarized in Table \ref{table:Overall}, our proposed DQ-DETR achieves the best result \textbf{30.2} AP compared with other state-of-the-art methods, including CNN-based and DETR-like methods. 
Also, DQ-DETR surpasses the baseline by 20.5\%, 20.6\%, 14.1\%, and 12.3\% in terms of $\text{AP}_{vt}$, $\text{AP}_{t}$, $\text{AP}_{s}$, $\text{AP}_{m}$. 
The performance gain is greater on $\text{AP}_{vt}$, and $\text{AP}_{t}$, and our DQ-DETR outperforms the advanced series of DETR-like models on AI-TOD-V2. We credit the performance gains for the following reasons: (1) DQ-DETR fuses the transformer visual features with a density map from the categorical counting module to improve the positional information of object queries, which makes the queries more suitable for localizing tiny objects. (2) Our dynamic query selection adaptively chooses an adequate number of object queries used for the detection task and can handle the images with either few or crowded objects.

\begin{table}[h!]
\caption{\textbf{Experiments on VisDrone.} All models are trained on the \textit{train} split and evaluated on the \textit{val} split. * denotes a re-implementation of the results.}
\centering
\resizebox{0.95\linewidth}{!}{
\begin{tblr}{
  width = \linewidth,
  colspec = {Q[438]Q[131]Q[158]Q[158]},
  column{even} = {c},
  column{3} = {c},
  vline{2} = {-}{},
  hline{1-2,9-10} = {-}{},
}
Model                        & AP   & $\text{AP}_{50}$ & $\text{AP}_{75}$ \\
Faster R-CNN~\cite{frcnn}    & 21.4 & 40.7 & 19.9 \\
Cascade R-CNN~\cite{crcnn}   & 22.6 & 38.8 & 23.2 \\
Yolov5~\cite{yolov5}         & 24.1 & 44.1 & 22.3 \\
CEASC~\cite{CEASC}           & 28.7 & 50.7 & 24.7 \\
SDP~\cite{SDP}               & 30.2 & 52.5 & 28.4 \\
DNTR~\cite{dntr}            & 33.1 & 53.8 & 34.8 \\ 
DINO-DETR*~\cite{DINO}       & 35.8 & 58.3 & 36.8 \\
\textbf{DQ-DETR (Ours)}               & \textbf{37.0} & \textbf{60.9} & \textbf{37.9} 
\end{tblr}
}
\label{table:visdrone}
\end{table}

\noindent \textbf{VisDrone.}
We also conduct experiments on the VisDrone~\cite{visdrone} dataset to demonstrate the effectiveness of our model DQ-DETR.
Table \ref{table:visdrone} shows our results on the VisDrone \textit{val} split. 
We compare the performances of our DQ-DETR with other methods.
Our proposed DQ-DETR achieves the best result \textbf{37.0} AP compared with other state-of-the-art methods, including CNN-based and DETR-like methods. 
Also, DQ-DETR surpasses the baseline DINO-DETR by 1.2, 2.6, and 1.1 in terms of AP, $\text{AP}_{50}$, $\text{AP}_{75}$. 

\noindent \textbf{COCO.}
\hspace{\parindent}We compare our method, DQ-DETR, with the previous state-of-the-art on the COCO dataset. DQ-DETR yielded slightly lower performance, with an AP of 50.2 compared to 51.3. We believe several factors contributed to these results. Firstly, our experiments were constrained by limited GPU resources, which may have impacted our ability to optimize the training process. Secondly, our method is specifically designed for tiny object detection in scenarios where the number of objects varies significantly across images. This specialized focus may not be fully leveraged in the COCO dataset, which is a general object detection task with a nearly balanced number of objects per image. Therefore, while our method shows potential, it may not perform as expected on general datasets like COCO due to various factors.

\begin{table}[h!]
\centering
\scriptsize
\caption{\textbf{Experiments on COCO.} All models are trained on the \textit{train} split and evaluated on the \textit{val} split.}
\begin{tblr}{
  width = \linewidth,
  colspec = {Q[295]Q[115]Q[79]Q[96]Q[96]Q[79]Q[83]Q[79]},
  column{even} = {c},
  column{3} = {c},
  column{5} = {c},
  column{7} = {c},
  vline{2-3} = {-}{},
  hline{1-2,8} = {-}{},
}
Method                                  & Epochs & AP & $\text{AP}_{50}$ & $\text{AP}_{75}$   & $\text{AP}_{S}$  & $\text{AP}_{M}$ & $\text{AP}_{L}$  \\
Faster R-CNN~\cite{frcnn}               & 109    & 42.0 & 62.1 & 45.5 & 26.6 & 45.5 & 53.4 \\
DETR~\cite{DETR}                        & 500    & 43.3 & 63.1 & 45.9 & 22.5 & 47.3 & 61.1 \\
Deformable DETR~\cite{Deformable-DETR}  & 50     & 46.2 & 65.2 & 50.0 & 28.8 & 49.2 & 61.7 \\
DN-DETR~\cite{DN-DETR}                  & 50     & 46.3 & 66.4 & 49.7 & 26.7 & 50.0 & 64.4 \\
DINO-DETR~\cite{DINO}                   & 24     & 51.3 & 69.1 & 56.0 & 34.5 & 54.2 & 65.8 \\
DQ-DETR                                 & 24     & 50.2 & 67.1 & 55.0 & 31.9 & 53.2 & 64.7 
\end{tblr}
\label{exp:COCO}
\end{table}
\vspace{-20pt}
\subsection{Ablation Study}
\hspace{\parindent}Categorical counting module, counting-guided feature enhancement, and dynamic query selection are the newly proposed contributions.
We conduct a series of ablation studies to verify the effectiveness of each component proposed in this paper.
DINO-DETR is chosen as the comparing DETR-like baseline.

\subsubsection{Main ablation experiment.}
\hspace{\parindent}Table \ref{ablation:Diff_module} shows the performance of our contributions separately on AI-TOD-V2. 
The results demonstrate that each component in DQ-DETR contributes to performance improvement. 
We attain an improved +2.2 AP over the baseline with the categorical counting module and dynamic query selection.
Furthermore, with feature enhancement refining the encoder's feature, it gains an extra improvement of +4.3, +2.6, +5.2 on AP, $\text{AP}_{vt}$, and $\text{AP}_{t}$ over the baseline.
Besides, the experiment with the counting module and feature enhancement together but without dynamic query selection further shows that introducing an additional counting-guided feature-enhancing task improves performance, even when query numbers remain static.
Consequently, we prove the power of each component in DQ-DETR on AI-TOD-V2.

\begin{table*}[h!]
\caption{Overall ablation for our architecture on AI-TOD-V2 \textit{test} split. Note that CC, DQS, and FE represent categorical counting, dynamic query selection, and feature enhancement, respectively.}
\centering
\resizebox{0.85\linewidth}{!}{
\scriptsize
\begin{tblr}{
  cells = {c},
  hline{1-2,6} = {-}{},
  vline{2-4} = {-}{},
}
{CC} & {DQS} & {FE} &AP   & $\text{AP}_{vt}$ & $\text{AP}_{t}$ & $\text{AP}_{s}$ & $\text{AP}_{m}$ \\
        &                          &                            &  25.9 & 12.7   & 25.3  & 32.0  & 39.7  \\
\checkmark & \checkmark                       &                 & 28.1 & 12.3   & 27.8  & 34.6  & 44.1  \\
\checkmark &                      &   \checkmark                & 29.1 & 14.4   & 29.3  & 35.2  & 44.1  \\
\checkmark & \checkmark        & \checkmark                     & \textbf{30.2} & \textbf{15.3}   & \textbf{30.5}  & \textbf{36.5}  & \textbf{44.6}  
\end{tblr}
}
\label{ablation:Diff_module}
\end{table*}
\vspace{-20pt}

\subsubsection{Ablation of DQ-DETR with different number of instances in images.}
\hspace{\parindent}We explore our DQ-DETR's performance under different numbers of instances in an image. 
We classify the AI-TOD-V2 dataset into 4 levels based on the number of instances $N$ in the image as in the categorical counting module, i.e., $N \leq 10$, $10 < N \leq 100$, $100 < N \leq 500$, and $500 < N$.
Our proposed DQ-DETR's performance is analyzed under these four situations. The results are shown in Table \ref{ablation:Diff_cls} compared with DINO-DETR as the baseline.
Our DQ-DETR dynamically adjusts the number of object queries based on the number of instances in the image, while DINO-DETR always uses 900 queries in all situations.

We can observe that in the situations of $N \leq 10$, and $10 < N \leq 100$, our DQ-DETR uses fewer numbers of queries and outperforms the baseline by 16\%, and 16.4\% in terms of AP. The performances in terms of $\text{AP}_{vt}$, $\text{AP}_{t}$ surpass the baseline by 19.8\%, and 20.8\% as well.
Moreover, it is noteworthy that DINO-DETR performs poorly when $N > 500$. Under these circumstances, there might be over 900 instances in some of the images, which is beyond the detection capability of DINO-DETR.
In dense images, the detection limitation of DINO-DETR with only 900 queries, leads to many objects undetected (FN), resulting in a lower AP.
Our DQ-DETR dynamically selects more queries for dense images, remarkably surpassing the baseline by 42.1\% in terms of $\text{AP}_{vt}$.

\begin{table*}[h!]
\caption{\textbf{Evaluation results for varying instance counts.} N indicates the number of instances in the image. We separate the AI-TOD-V2 dataset into 4 classes based on N. All models are trained on the AI-TOD-V2 \textit{trainval} split and evaluated on \textit{test} split.}
\centering
\resizebox{0.98\linewidth}{!}{
\begin{tblr}{
  column{3} = {c},
  column{4} = {c},
  column{5} = {c},
  column{6} = {c},
  column{7} = {c},
  column{8} = {c},
  column{9} = {c},
  column{10} = {c},
  vline{2-4} = {-}{},
  hline{1-2,7,12} = {-}{},
}
Model       & \#Objects in image         & \#Query & AP   & $\text{AP}_{50}$ & $\text{AP}_{75}$ & $\text{AP}_{vt}$ & $\text{AP}_{t}$ & $\text{AP}_{s}$ & $\text{AP}_{m}$ \\
            & N $\leq 10$                 & 900     & 22.5 & 53.1 & 14.8 & 10.6   & 24.5  & 25.7  & 34.9  \\
            & $10 <$ N $\leq 100$         & 900     & 24.4 & 58.8 & 15.9 & 13.0   & 22.9  & 31.5  & 37.3  \\
DINO-DETR~\cite{DINO}   & $100 <$ N $\leq 500$        & 900     & 31.6 & 67.3 & 26.9 & 10.1   & 25.4  & 39.6  & 38.2  \\
            & $500 <$ N                & 900     & 13.5 & 27.9 & 7.3  & 5.7    & 6.4   & 34.7  & 32.4  \\
            & Overall                   & 900     & 25.9 & 61.3 & 17.5 & 12.7   & 25.3  & 32.0  & 39.7   \\
\hline
            & N $\leq 10$                 & 300     & 26.1 & 60.4 & 19.7 & 12.7   & 29.6  & 28.5  & 40.8  \\
            & $10 <$ N $\leq 100$         & 500     & 28.4 & 65.9 & 20.1 & 15.2   & 27.8  & 34.7  & 41.8  \\
DQ-DETR (Ours)   & $100 <$ N $\leq 500$    & 900     & 33.7 & 69.9 & 30.4 & 11.1   & 30.4  & 42.0  & 41.6  \\
            & $500 <$ N                & 1500    & 14.7 & 35.6 & 7.5  & 8.1    & 7.8   & 37.5  & 40.4  \\
            & Overall                   & Dynamic & 30.2 & 68.6 & 22.3 & 15.3   & 30.5  & 36.5  & 44.6
\end{tblr}
}
\label{ablation:Diff_cls}
\end{table*}
\subsubsection{Ablation of Categorical Counting Module.}
\hspace{\parindent}Table \ref{ablation:cls_acc} demonstrates the accuracy of the classification task in our categorical counting module. 
The performance is analyzed under four situations, where $N$ is the number of instances per image.
The total classification accuracy is about 94.6\%, which means our categorical counting module can accurately estimate the number of objects $N$ in the images.
Furthermore, we can find that our categorical counting module has a poor classification performance with only 56.6\% accuracy in the $N > 500$ situation since the number of training images is much fewer in this situation. 
Also, we observe that there are at most 2267 instances per image in the AI-TOD-V2 dataset. However, the long-tailed distribution of the training samples restricts us from classifying the number of instances $N$ in more detail. We have no choice but to categorize the images with $500 < N \leq 2267$ into the same class.

As for the detection accuracy, our DQ-DETR outperforms the baseline under all situations. 
The performances surpass the baseline by 16\% and 16.4\% in terms of AP for $N \leq 10$, $10 < N \leq 100$.
Nevertheless, our DQ-DETR performs slightly better than the baseline in the scenario with $N > 500$.  
This phenomenon is due to the poor classification accuracy for $N > 500$.
The incorrect prediction from the categorical counting module directly affects the number of object queries used for detection, where the inappropriate number of queries might harm the detection performance.

\begin{table*}[h!]
\caption{The classification accuracy of our categorical counting module and detection accuracy of DQ-DETR with different numbers of instances in the images. DINO-DETR is compared as the baseline.}
\centering
\resizebox{0.98\linewidth}{!}{
\begin{tblr}{
  column{even} = {c},
  column{3} = {c},
  column{5} = {c},
  vline{2-5} = {-}{},
  hline{1-2,6-7} = {-}{},
}
\#Objects in image                & Accuracy(\%) & AP (DQ-DETR) & AP (Baseline) & \#Sample \\
N $\leq 10$           & 97.7         & 26.1 (+3.6)      & 22.5          & 8674     \\
$10 <$ N $\leq 100$   & 90.5         & 28.4 (+4.0)      & 24.4          & 4393     \\
$100 <$ N $\leq 500$  & 86.5         & 33.7 (+2.1)      & 31.6          & 905      \\
$500 <$ N          & 56.5         & 14.7 (+1.2)      & 13.5          & 46       \\
Total               & 94.6         & 30.2             & 25.9          & 14018    
\end{tblr}
}
\label{ablation:cls_acc}
\end{table*}

\begin{table}[h!]
\caption{Ablation of using regression or classification in categorical counting module.}
\centering
\begin{tblr}{
  column{2} = {c},
  column{3} = {c},
  column{4} = {c},
  column{5} = {c},
  column{6} = {c},
  vline{2} = {-}{},
  hline{1-2,5} = {-}{},
}
Method         & AP   & $\text{AP}_{vt}$ & $\text{AP}_{t}$ & $\text{AP}_{s}$ & $\text{AP}_{m}$ \\
Baseline       & 25.9 & 12.7   & 25.3  & 32.0  & 39.7  \\
Regression     & 14.9 & 5.2    & 16.3  & 19.9  & 14.3  \\
Classification & 30.2 & 15.3   & 30.5  & 36.5  & 44.6  
\end{tblr}
\label{ablation:Cls_Reg}
\end{table}

Table \ref{ablation:Cls_Reg} compares our DQ-DETR's performance of using classification or regression in the categorical counting module.
The traditional crowd-counting methods usually regress the predicted counting number to a specific value.
However, in our study, we use a classification head instead. This experiment demonstrates our DQ-DETR performance with these two methods. 
For the classification task, we classify the images into 4 classes and apply different numbers of queries in the transformer decoder, as we mentioned in the previous section.
For the regression task, we regress an integer directly to predict the number of objects in the image and select the object queries corresponding to the predicted result.

The results demonstrate that using regression as a counting method performs extremely poorly.
We impute the drastic performance drop for the following reasons: (1) It is challenging to regress an accurate number since the number of instances per image may vary significantly from 1 to 2267 in the AI-TOD-V2 dataset. (2) Unstable regression results significantly affect the number of queries used in the transformer decoder, making it difficult for the DETR model to converge. Owing to the above reasons, we believe that classifying how many objects exist in the image into different levels is a simpler way in contrast to regression. Thus, classification instead of regression is preferred as a method in our proposed categorical counting module.

\section{Conclusion}
\label{sec:conclusion}
\hspace{\parindent}In this paper, we analyze that the fixed number and position of queries in previous DETR-like methods are unsuitable for detecting tiny objects in aerial datasets and propose a new end-to-end transformer detector DQ-DETR with a categorical counting module, counting-guided feature enhancement, and dynamic query selection.
Our DQ-DETR dynamically adjusts the number of object queries used for detection to solve the imbalance of instances between different aerial images.
Also, we improve the positional information of queries, making it easier for the decoder to localize the tiny object.
DQ-DETR is the first DETR-like model focusing on tiny object detection and achieves 30.2\% $\mathrm{AP}$, which is the state-of-the-art of AI-TOD-V2.
The result shows that our proposed DQ-DETR improves the performance of detecting tiny objects, outperforming all previous CNN-based detectors and DETR-like methods on the AI-TOD-V2 dataset with ResNet50 as the backbone. 

\section*{Acknowledgment}
This work is partially supported by the National Science and Technology Council, Taiwan under Grants NSTC-112-2221-E-A49-059-MY3, NSTC-112-2221-E-A49-094-MY3, NSTC-112-2628-E-002-033-MY4 and NSTC-112-2634-F-002-002-MBK, and was financially supported in part by the Center of Data Intelligence: Technologies, Applications, and Systems, National Taiwan University (Grants: 113L900901/113L900902/113L900903), from the Featured Areas Research Center Program within the framework of the Higher Education Sprout Project by the Ministry of Education, Taiwan.

\bibliographystyle{splncs04}
\bibliography{main}

\begin{thebibliography}{10}
\providecommand{\url}[1]{\texttt{#1}}
\providecommand{\urlprefix}{URL }
\providecommand{\doi}[1]{https://doi.org/#1}

\bibitem{crcnn}
Cai, Z., Vasconcelos, N.: Cascade r-cnn: Delving into high quality object detection. In: CVPR (2018)

\bibitem{DETR}
Carion, N., Massa, F., Synnaeve, G., Usunier, N., Kirillov, A., Zagoruyko, S.: End-to-end object detection with transformers. In: ECCV (2020)

\bibitem{Dynamic_DETR}
Dai, X., Chen, Y., Yang, J., Zhang, P., Yuan, L., Zhang, L.: Dynamic detr: End-to-end object detection with dynamic attention. In: ICCV (2021)

\bibitem{CEASC}
Du, B., Huang, Y., Chen, J., Huang, D.: Adaptive sparse convolutional networks with global context enhancement for faster object detection on drone images. In: CVPR (2023)

\bibitem{yolov5}
Jocher, G.: Yolov5 by ultralytics. \url{https://github.com/ultralytics/yolov5} (2023), accessed: 2024-07-08

\bibitem{da_sm}
Kisantal, M., Wojna, Z., Murawski, J., Naruniec, J., Cho, K.: Augmentation for small object detection. arXiv preprint arXiv:1902.07296  (2019)

\bibitem{DN-DETR}
Li, F., Zhang, H., Liu, S., Guo, J., Ni, L.M., Zhang, L.: Dn-detr: Accelerate detr training by introducing query denoising. In: CVPR (2022)

\bibitem{FPN}
Lin, T.Y., Dollár, P., Girshick, R.B., He, K., Hariharan, B., Belongie, S.J.: Feature pyramid networks for object detection. In: CVPR (2017)

\bibitem{focal_loss}
Lin, T.Y., Goyal, P., Girshick, R., He, K., Dollar, P.: Focal loss for dense object detection. TPAMI  \textbf{42}(2),  318--327 (2020)

\bibitem{dntr}
Liu, H.I., Tseng, Y.W., Chang, K.C., Wang, P.J., Shuai, H.H., Cheng, W.H.: A denoising fpn with transformer r-cnn for tiny object detection. TGRS  \textbf{62},  1--15 (2024)

\bibitem{DAB-DETR}
Liu, S., Li, F., Zhang, H., Yang, X., Qi, X., Su, H., Zhu, J., Zhang, L.: {DAB}-{DETR}: Dynamic anchor boxes are better queries for {DETR}. In: ICLR (2022)

\bibitem{SDP}
Ma, Y., Chai, L., Jin, L.: Scale decoupled pyramid for object detection in aerial images. TGRS  \textbf{61},  1--14 (2023)

\bibitem{Conditional_DETR}
Meng, D., Chen, X., Fan, Z., Zeng, G., Li, H., Yuan, Y., Sun, L., Wang, J.: Conditional detr for fast training convergence. In: ICCV (2021)

\bibitem{LRP_ECCV}
Oksuz, K., Cam, B.C., Akbas, E., Kalkan, S.: Localization recall precision (lrp): A new performance metric for object detection. In: ECCV. pp. 504--519 (2018)

\bibitem{detectors}
Qiao, S., Chen, L.C., Yuille, A.: Detectors: Detecting objects with recursive feature pyramid and switchable atrous convolution. In: CVPR (2021)

\bibitem{yolov3}
Redmon, J., Farhadi, A.: Yolov3: An incremental improvement. arXiv preprint arXiv:1804.02767  (2018)

\bibitem{frcnn}
Ren, S., He, K., Girshick, R., Sun, J.: Faster r-cnn: Towards real-time object detection with region proposal networks. Advances in neural information processing systems  \textbf{28} (2015)

\bibitem{GIOU}
Rezatofighi, S.H., Tsoi, N., Gwak, J., Sadeghian, A., Reid, I.D., Savarese, S.: Generalized intersection over union: A metric and a loss for bounding box regression. In: CVPR (2019)

\bibitem{Sun_2021_ICCV}
Sun, Z., Cao, S., Yang, Y., Kitani, K.M.: Rethinking transformer-based set prediction for object detection. In: ICCV (2021)

\bibitem{fcos}
Tian, Z., Shen, C., Chen, H., He, T.: Fcos: Fully convolutional one-stage object detection. In: ICCV (2019)

\bibitem{nwd}
Wang, J., Xu, C., Yang, W., Yu, L.: A normalized gaussian wasserstein distance for tiny object detection. arXiv preprint arXiv:2110.13389  (2021)

\bibitem{AI-TOD}
Wang, J., Yang, W., Guo, H., Zhang, R., Xia, G.S.: Tiny object detection in aerial images. In: ICPR (2021)

\bibitem{Anchor_DETR}
Wang, Y., Zhang, X., Yang, T., Sun, J.: Anchor detr: Query design for transformer-based detector. In: AAAI (2022)

\bibitem{CBAM}
Woo, S., Park, J., Lee, J.Y., Kweon, I.S.: Cbam: Convolutional block attention module. In: ECCV (2018)

\bibitem{AITODv2}
Xu, C., Wang, J., Yang, W., Yu, H., Yu, L., Xia, G.S.: Detecting tiny objects in aerial images: A normalized wasserstein distance and a new benchmark. ISPRS  \textbf{190},  79--93 (2022)

\bibitem{rfla}
Xu, C., Wang, J., Yang, W., Yu, H., Yu, L., Xia, G.S.: Rfla: Gaussian receptive field based label assignment for tiny object detection. In: ECCV (2022)

\bibitem{dot}
Xu, C., Wang, J., Yang, W., Yu, L.: Dot distance for tiny object detection in aerial images. In: CVPRW (2021)

\bibitem{DINO}
Zhang, H., Li, F., Liu, S., Zhang, L., Su, H., Zhu, J., Ni, L.M., Shum, H.Y.: Dino: Detr with improved denoising anchor boxes for end-to-end object detection. In: ICLR (2023)

\bibitem{DDQ}
Zhang, S., Wang, X., Wang, J., Pang, J., Lyu, C., Zhang, W., Luo, P., Chen, K.: Dense distinct query for end-to-end object detection. In: CVPR (2023)

\bibitem{visdrone}
Zhu, P., Wen, L., Du, D., Bian, X., Fan, H., Hu, Q., Ling, H.: Detection and tracking meet drones challenge. TPAMI  \textbf{44}(11),  7380--7399 (2021)

\bibitem{Deformable-DETR}
Zhu, X., Su, W., Lu, L., Li, B., Wang, X., Dai, J.: Deformable detr: Deformable transformers for end-to-end object detection. In: ICLR (2021)

\bibitem{learn_da}
Zoph, B., Cubuk, E.D., Ghiasi, G., Lin, T.Y., Shlens, J., Le, Q.V.: Learning data augmentation strategies for object detection. In: ECCV (2020)

\end{thebibliography}

\renewcommand\thefigure{\Alph{section}\arabic{figure}}
\renewcommand\thetable{\Alph{section}\arabic{table}}
\setcounter{figure}{0}
\setcounter{table}{0}
\renewcommand{\thesection}{\Alph{section}}
\setcounter{section}{0}

\appendix
\title{Appendix}

\author{}
\institute{}

\maketitle

\vspace{-3em}

\section{Quantitative Results}

\subsection{FP/FN Under Different Density Situation}
\hspace{\parindent}In Table \ref{ablation:LRP}, we explore our DQ-DETR's performance under different density situations, including sparse and dense images. 
We classify images with less than 100 objects as sparse and images with over 900 objects as dense.
LRP FP and LRP FN~\cite{LRP_ECCV} are used as the evaluation metric.
Unlike AP metrics, a lower LRP value implies better performance.
Previous DETR-like models apply a fixed number of object queries, while our DQ-DETR uses a dynamic number of queries depending on the object’s density in the picture.

DINO-DETR uses a fixed number of 900 queries for detection, no matter whether in dense or sparse situations. The number of queries exceeds the number of objects in the sparse image and hence introduces many underlying false positive samples (FP).
In contrast, for dense images, the number of queries DINO-DETR uses is far less than the number of objects in images, which is beyond the detection capability of DINO-DETR, leaving lots of instances undetected (FN) and causes a large LRP FN score.
Our proposed DQ-DETR dynamically adjusts the number of queries used for detection, resulting in fewer FP in sparse images and fewer FN in dense images.
\vspace{-10pt}
\begin{table}[h!]
\centering
\caption{LRP FP and LRP FN score under different density situations in AI-TOD-V2. DINO-DETR is compared as the baseline.}
\begin{tblr}{
  width = \linewidth,
  column{3} = {c},
  column{4} = {c},
  column{2} = {c},
  cell{2}{1} = {r=2}{},
  cell{4}{1} = {r=2}{},
  vline{2} = {1-5}{},
  vline{3} = {1-5}{},
  hline{1-2} = {-}{},
  hline{4,6} = {-}{},
}
Method    & Situation & LRP FP & LRP FN \\
DINO-DETR & Sparse    & 29.4   & 40.7   \\
          & Dense     & 36.8   & 75.1   \\
DQ-DETR   & Sparse    & 25.7   & 36.4   \\
          & Dense     & 35.4   & 51.5   
\end{tblr}
\label{ablation:LRP}
\end{table}
\vspace{-20pt}
\subsection{Ablation of Categorical Counting Module}

\hspace{\parindent}In the categorical counting module, we categorize the number of objects $N$ per image into four levels, which are $N \leq 10$, $10 < N \leq  100$, $100 < N \leq 500$, and $N > 500$.
We selected the numbers 10, 100, and 500 based on the AI-TOD-V2 dataset's characteristics, i.e., the mean and standard deviation of the number of instances $N$ per image.
Further, we only classify the number of objects $N$ into four levels due to the long-tail distribution of the training samples.
For the $N > 500$ situation, there are only 46 training images in this situation, which is much fewer than other cases and leads to a poor classification performance with only 56.6\% accuracy.
Although there are at most 2267 instances per image in the AI-TOD-V2 dataset, the long-tail distribution of the training samples restricts us from classifying the number of instances $N$ per image in a more detailed manner.

Table \ref{ablation:num_cls} and Table \ref{ablation:5class} demonstrate the detection performance and the accuracy of the classification task in our categorical counting module.
We can observe that if we classify the number of objects $N$ into more classes, e.g., 5 classes, AP drops 1.4 compared to the 4-class scenario.
That is because the poor classification results from the categorical counting module will directly affect the number of object queries used for detection and the inappropriate number of queries might harm the detection performance.
In the 5-class classification scenario, while the total classification accuracy maintains 93.8\%, the accuracy in the $500 <$ N $\leq 900$, and $N > 900$ situations are only 37.1\% and 57.4\%.
Since there are only a few training images, the categorical counting module doesn't perform well in these two situations and further impacts the detection performance.
Hence, we only categorize the number of objects $N$ per image into four levels without partitioning the $N > 500$ situation into more detailed settings.
\vspace{-15pt}

\begin{table}[h!]
\centering
\caption{Ablation of categorical counting module. DINO-DETR is compared as the baseline.}
\resizebox{0.7\linewidth}{!}{
\begin{tblr}{
  column{even} = {c},
  column{3} = {c},
  column{5} = {c},
  vline{2} = {-}{},
  hline{1-2,5} = {-}{},
}
Method               & AP   & $\text{AP}_{vt}$ & $\text{AP}_{t}$ & $\text{AP}_{s}$ & $\text{AP}_{m}$ \\
Baseline             & 25.9 & 12.7   & 25.3  & 32.0  & 39.7  \\
Classification (4cls) & \textbf{30.2} & \textbf{15.3}   & \textbf{30.5}  & \textbf{36.5}  & \textbf{44.6}  \\
Classification (5cls) & 28.8 & 14.3   & 29.2  & 34.1  & 43.1  
\end{tblr}
}
\label{ablation:num_cls}
\end{table}

\begin{table}[h!]
\centering
\caption{The classification accuracy of our categorical counting module with different numbers of classes.}
\resizebox{0.9\linewidth}{!}{
\begin{tblr}{
  width = \linewidth,
  column{even} = {c},
  column{3} = {c},
  vline{2-4} = {-}{},
  hline{1-2,7-8} = {-}{},
}
\#Objects in image           & Accuracy(\%) @ 4cls & Accuracy(\%) @ 5cls & \#Sample \\
N $\leq 10$                  & 97.7                & 97.5                & 8674     \\
$10 <$ N $\leq 100$          & 90.5                & 89.3                & 4393     \\
$100 <$ N $\leq 500$         & 86.5                & 83.2                & 905      \\
$500 <$ N $\leq 900$         & 56.5                & 37.1                & 35       \\
$900 <$ N                    & -                   & 54.4                & 11       \\
Total                        & 94.6                & 93.8                & 14018    
\end{tblr}
}
\label{ablation:5class}
\end{table}

\subsection{Categorical Counting Module (CCM) for Different Datasets.}
Since the characteristics of different datasets may vary a lot, it is vital to tailor our categorical counting module to the dataset property and determine how many queries should be used for object detection. 
In the categorical counting module for AI-TOD-V2, we estimate the counting number $N$, i.e., the number of instances per image, by a classification head and categorize them into four levels, which are $N \leq 10$, $10 < N \leq  100$, $100 < N \leq 500$, and $N > 500$. It is worth noting that the hyperparameters 10, 100, and 500 are tailored for AI-TOD-V2. For other datasets, we recommend adjusting the hyperparameters used in the CCM through a logical process that considers the mean and variance of the objects per image in the dataset. 
This approach can reduce the need to manually design the CCM for different datasets.


\begin{algorithm}
\caption{Pseudo Code for Categorical Counting Module}
\begin{algorithmic}[1]
\Function{Categorical-Counting}{$features$}
    \State $mean \gets \text{Mean}(dataset)$
    \State $var \gets \text{Variance}(dataset)$
    
    \State $GT \gets \text{Predict}(features)$
    \State $class1 \gets \{GT < mean - var \}$
    \State $class2 \gets \{mean - var \leq GT < mean\}$
    \State $class3 \gets \{mean \leq GT < mean + var\}$
    \State $class4 \gets \{GT \geq mean + var\}$
    
\EndFunction
\end{algorithmic}
\end{algorithm}

\renewcommand{\thefigure}{B1}
\begin{figure}[ht!]
    \centering
    \includegraphics[width=0.7\linewidth]{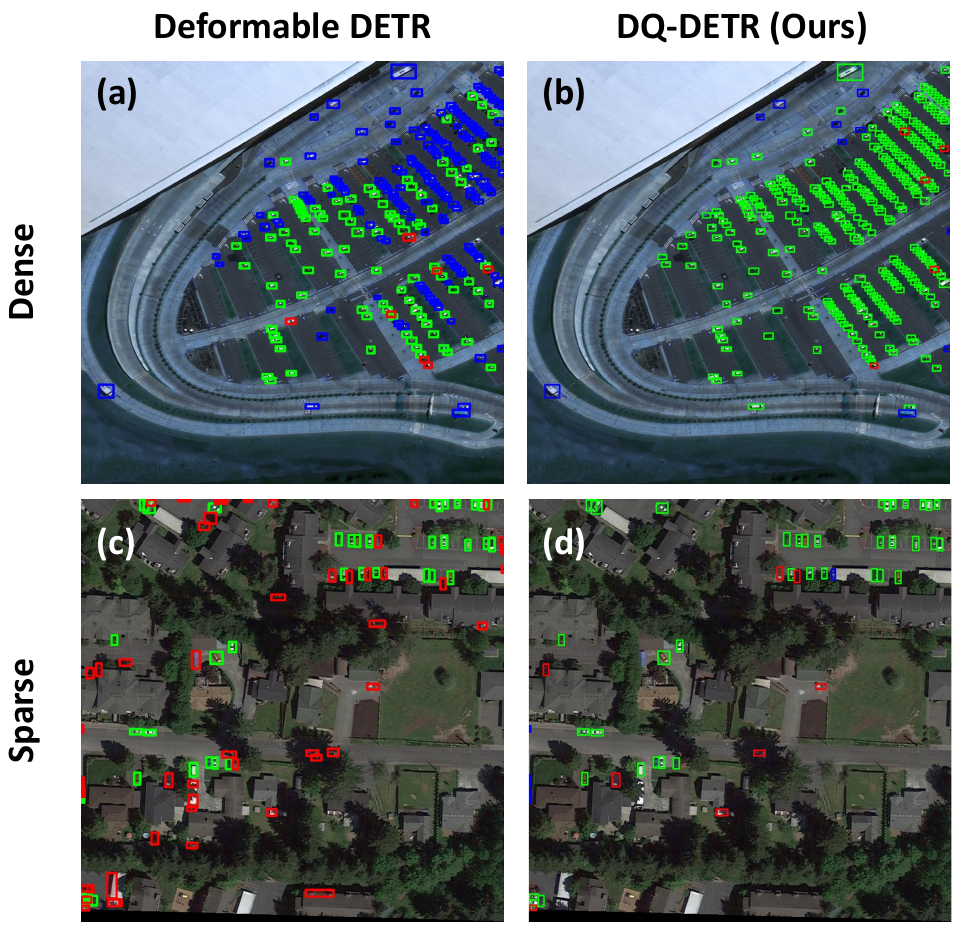}
    \caption{Visualization of detection results and feature maps. The green, red, and blue boxes represent TP, FP, and FN, respectively.}
    \label{fig:visual}
\end{figure}

\vspace{-20pt}
\section{Visualization}
Fig. \ref{fig:visual} presents the detection results of our DQ-DETR alongside Deformable-DETR. By selecting an appropriate number of object queries, DQ-DETR effectively detects most of the tiny objects in dense scenes, demonstrating superior performance in capturing fine details. Conversely, Deformable-DETR suffers from an insufficient number of object queries, leading to a higher rate of undetected objects (false negatives). In sparse scenes, Deformable-DETR further struggles due to the use of excessive object queries with unrefined positional information, resulting in an increased number of false positives in the detection outcomes.


\end{document}